\documentclass{article}

\usepackage{iclr2027_conference,times}

\newif\ifshowappendix
\showappendixtrue

\usepackage[utf8]{inputenc} 
\usepackage[T1]{fontenc}    
\usepackage{hyperref}       
\hypersetup{hidelinks}
\usepackage{url}            
\usepackage{booktabs}       
\usepackage{amsfonts}       
\usepackage{nicefrac}       
\usepackage{microtype}      
\usepackage{xcolor}         
\usepackage{graphicx}
\usepackage{makecell}
\usepackage{multirow}
\usepackage{tabularx}
\usepackage{amsmath}
\usepackage{subcaption}
\usepackage{enumitem}
\usepackage{chngcntr}
\usepackage{placeins}

\title{When Normal-Only Updates Are Not Enough: Test-Time Adaptation in Reconstruction-Based Time-Series Anomaly Detection}

\author{%
  Jianhuan Mao \\
  Beihang University \\
  \And
  Mengxiao Zhu \\
  North China University of Technology \\
  \And
  Yue Pei \\
  Beihang University \\
  \And
   Lei Li \\
  Zhongguancun Laboratory \\
  \And
  Haogang Zhu \\
  Beihang University \\
}
\iclrfinalcopy

\begin{document}
\maketitle
\lhead{Preprint}

\begin{abstract}
Normal operating conditions can evolve after deployment, motivating test-time adaptation (TTA) of time-series anomaly detectors using incoming data. 
Yet the same stream may contain anomalies, so TTA with online updates must decide which test samples should be used for adaptation. 
This leaves a more basic question: even if updates are restricted to samples labeled as normal in the benchmark, does TTA reliably improve anomaly detection?
We study this question with a controlled intervention that replaces the practical update selector with a rule that uses only samples labeled as normal in the benchmark, while keeping the rest of the online adaptation procedure unchanged. 
Across nine benchmark families, four optimizer configurations, and five random seeds, we obtain three main findings: (1) using only samples labeled as normal in the benchmark does not reliably improve detection; (2) practical TTA failures exhibit two distinct responses to this intervention, with 63 of 180 settings recovering a clear detection gain while 90 still show no clear detection gain; and (3) the updates reduce their immediate reconstruction objective across the evaluated settings, yet these reductions do not consistently lead to better anomaly ranking. Both failure patterns persist across alternative practical adaptation variants and modest perturbations of the benchmark labels. Together, these findings show that improving sample selection alone is not a complete solution to reconstruction-based TSAD TTA.
\end{abstract}

\section{Introduction}
Time series anomaly detection (TSAD) is widely used to monitor complex systems and detect departures from normal operating behavior~\citep{IoT-survey, RobustTSA}. Many TSAD detectors learn or calibrate a notion of normality from historical observations, yet the normal operating conditions encountered after deployment may continue to evolve~\citep{review_TSAD}. Under distribution shift after deployment, a fixed detector may assign high anomaly scores to shifted but benign observations, producing false positives~\citep{M2N2,CANDI}. This deployment setting motivates test-time adaptation (TTA), which updates a deployed detector during inference using incoming test data, allowing adaptation without waiting for a separate offline refitting cycle~\citep{TENT,OTTA_survey,M2N2,CANDI}.

However, the same test stream that enables adaptation also creates a basic challenge. 
It contains both evolving normal observations that may support adaptation and the anomalies the detector is expected to detect. 
Indiscriminate updates therefore risk adapting the detector to anomalous observations rather than benign changes in normal behavior. 
Recent TTA methods for reconstruction-based TSAD therefore make explicit choices about which test samples drive adaptation, and some also constrain how the detector is updated.
For example, M2N2~\citep{M2N2} updates the detector using samples predicted as normal, whereas CANDI~\citep{CANDI} further curates informative candidates using anomaly scores and similarity in the latent space and restricts adaptation to a lightweight residual module.
These designs show that careful sample selection addresses one important source of risk, but does not by itself determine whether the resulting updates improve anomaly detection.
This leaves a basic question unresolved: if test-time updates are restricted to samples labeled as normal in the benchmark, would adaptation reliably improve anomaly detection?

To answer this question, we construct an oracle setting using benchmark labels.
In this setting, the practical selection rule is replaced by one that uses only samples labeled as normal in the benchmark, while keeping the source initialization, updater, optimizer configuration, and online protocol unchanged.
Our first finding is that adaptation using only samples labeled as normal in the benchmark can lead to very different detection outcomes. Across nine benchmark families, four optimizer configurations, and five random seeds, it produces substantial gains in some settings, little change in others, and degradation in some cases.

Our second finding is that practical TTA failures do not all respond to this intervention in the same way.
Under our primary practical selector, which selects samples predicted as normal using a score threshold, 63 of the 180 matched cells show no clear detection gain under practical TTA but recover a clear gain when updates are restricted to samples labeled as normal in the benchmark.
In contrast, 90 cells still show no clear detection gain even after the same replacement.
Both patterns persist across alternative practical adaptation variants and modest perturbations of the benchmark labels.
Taken together, these results show that changing sample selection can yield clear detection gains in some cases where practical TTA fails, while leaving others unresolved, motivating a closer examination of the adaptation behavior in the latter cases.

Failures that remain after restricting updates to samples labeled as normal in the benchmark raise a deeper question: could they simply reflect a failure to improve the reconstruction objective used for adaptation?
Our third finding is that improving the reconstruction objective does not reliably imply better anomaly detection: the updates reduce their immediate reconstruction objective across the evaluated cells, while detection gains vary substantially.
An analysis across all 180 cells further shows that larger detection gains are associated with more favorable correction of difficult normal--anomaly score orderings. 
This association is driven primarily by differences across datasets and adaptation settings, rather than by a consistent relationship across optimizers or seeds within individual datasets.

Together, our study yields three main empirical insights into reconstruction-based TSAD TTA.
(1) Benchmark-labeled normal-only updates do not guarantee improved anomaly detection. Even when updates are restricted to samples labeled as normal in the benchmark, detection can improve substantially, change little, or degrade.
(2) Practical TTA failures are not homogeneous. Some failures are rescued when practical selection is replaced by benchmark-labeled normal-only selection, whereas others persist under the same intervention. Thus, changing sample selection alone does not resolve all practical-TTA failures.
(3) Improving the reconstruction objective does not reliably imply improved anomaly ranking. Across datasets and adaptation settings, larger detection gains are associated with more favorable changes in the relative scores of difficult normal and anomalous samples, although this relationship is primarily a between-dataset/regime pattern.
To establish these insights, we use a matched label-conditioned intervention that changes update-sample selection while keeping the remaining online adaptation configuration fixed.

\section{Related Work}

\paragraph{TSAD under distribution shift and test-time adaptation.}
Most TSAD methods are trained under a relatively stable notion of normality, but deployment-time drift can cause shifted yet benign patterns to receive high anomaly scores. Reconstruction-based detectors are a common TSAD family, ranging from autoencoders~\citep{AE} to recurrent encoder--decoders~\citep{LSTMED} and adversarial reconstruction models such as USAD~\citep{USAD}. Because reconstruction error naturally provides both an anomaly score and an update objective, reconstruction-based detectors also provide a direct basis for test-time model adaptation. Earlier streaming anomaly-detection work considered adaptation under drift, including prediction-error-based incremental RNN adaptation and FITNESS-style fine-tuning on recent similar samples in streams with drift and outliers~\citep{RNN-AD,Fitness}. More recent TSAD-TTA methods use explicit test-time update protocols with reconstruction-based detectors. M2N2 updates detector parameters using samples predicted as normal after trend-based normalization~\citep{M2N2}, whereas CANDI curates informative adaptation samples using anomaly scores and latent-space similarity and restricts updates to a lightweight residual module~\citep{CANDI}. Despite their different selection and update designs, both methods perform reconstruction-based test-time adaptation using selected test samples. Our work is complementary: rather than proposing another selector or update rule, we isolate the role of update-sample selection by replacing the evaluated selector with benchmark-labeled normal-only selection under a matched adaptation configuration, and study when this intervention does and does not recover detection gains.

\paragraph{General test-time adaptation.}
Test-time adaptation (TTA) addresses distribution shift by adapting a trained model using unlabeled test data during inference. Prior online TTA methods, including TENT, CoTTA, and EATA, show that adaptation behavior can depend strongly on sample filtering, optimization stability, continual updating, and evaluation protocol~\citep{TENT,COTTA,EATA,OTTA_survey}. These issues are particularly relevant to TSAD because the same incoming stream supplies adaptation data while also containing the anomalies to be detected. Consequently, selecting which test samples drive updates is intertwined with how online parameter updates reshape the detector. Our study focuses on this interaction in reconstruction-based TSAD, where the update-sample selector can be changed while the remaining adaptation configuration is held fixed.

\paragraph{Benchmarking and evaluation in TSAD.}
Recent TSAD benchmark studies show that empirical conclusions can be sensitive to dataset choice, post-processing, windowing protocols, evaluation metrics, and benchmark-label quality. TAB documents inconsistencies in dataset splitting, threshold selection, point adjustment, and output protocols~\citep{TAB}, while TSB-AD highlights flaws in existing datasets and evaluation measures and advocates more reliable benchmarking practices~\citep{TSB-AD}. These observations are particularly relevant to our label-conditioned intervention: benchmark labels provide a controlled way to define normal-only update samples, but they should not be interpreted as a perfect semantic notion of normality. Our work does not introduce a new benchmark; instead, it uses a matched intervention across existing TSAD benchmarks to study whether replacing practical update selection with benchmark-labeled normal-only selection is sufficient to recover detection gains, and how adaptation behavior relates to the downstream anomaly-ranking objective.
\section{Problem Setup and Matched Intervention}
\label{sec:setup}

We consider reconstruction-based TSAD with online test-time parameter updates. Let $X_t=(x_{t,1},\ldots,x_{t,w})$ denote the test window processed at online step $t$, viewed as an element of $\mathbb{R}^{w\times D}$ by stacking the $w$ observations along the time dimension, where $w$ is the window length and $D$ is the number of variables. Here, $x_{t,i}\in\mathbb{R}^{D}$ is the multivariate observation at position $i\in\{1,\ldots,w\}$ within the window. Each point has a label $y_{t,i}\in\{0,1\}$, where $0$ denotes normal and $1$ denotes anomalous. 
A reconstruction detector with parameters $\theta$ outputs
$\hat{x}_{t,i}(\theta)\in\mathbb{R}^D$ and assigns the pointwise anomaly
score
\[
s_{t,i}(\theta)
=
\ell(x_{t,i},\hat{x}_{t,i}(\theta)),
\]
where $\ell$ denotes a scalar pointwise reconstruction loss. In the main autoencoder (AE) experiments, we instantiate it as the mean squared reconstruction error over the $D$ variables,
\[
\ell(x_{t,i},\hat{x}_{t,i}(\theta))
=
\frac{1}{D}
\sum_{d=1}^{D}
\left(
\hat{x}_{t,i,d}(\theta)-x_{t,i,d}
\right)^2.
\]
This pointwise loss is used directly as the anomaly score, with larger values
indicating higher anomaly likelihood. Detection is evaluated by ranking test points according to these scalar scores.

Test-time adaptation updates the detector using selected test points. We represent this selection by a point-level binary mask $m_t=(m_{t,1},\ldots,m_{t,w})\in\{0,1\}^{w}$, where $m_{t,i}=1$ means that position $i$ in window $X_t$ contributes to the update loss. The mask selects time points rather than individual variables. Given $m_t$, the test-time reconstruction objective is
\[
\mathcal{L}_{\mathrm{TTA}}(\theta;X_t,m_t)
=
\frac{1}{w}
\sum_{i=1}^{w}
m_{t,i}\,
\ell(x_{t,i},\hat{x}_{t,i}(\theta)).
\]
Unselected points contribute zero loss but remain in the window average; equivalently, point losses are zero-filled by the mask before averaging over the full window. The selector therefore determines which points contribute to the update objective and how much selected data contributes to the averaged update loss.

We compare three matched settings that differ in whether and how test points are selected for online updates. The frozen detector keeps the source parameters fixed, $\theta^{F}_t=\theta_0$ for all $t$. 
Practical TTA, denoted by $P$, uses our primary score-thresholded predicted-normal selector,
\[
m^P_{t,i}
=
\mathbf{1}\{s_{t,i}(\theta_t^P) < \tau_{\mathrm{sel}}\},
\]
where $\tau_{\mathrm{sel}}$ is calibrated separately for each physical entity
and seed as the 95th percentile of point-level reconstruction scores produced
by the frozen source checkpoint on the chronological source-validation split.
The threshold is fixed throughout online testing and is never recalibrated
using test data. This selector serves as the primary deployable baseline for
the matched comparison and is not intended to represent all possible practical
selection strategies. Oracle normal-only TTA, denoted by $O$, replaces this mask with the benchmark-label-conditioned normal mask,
\[
m^{O}_{t,i}=\mathbf{1}\!\left\{y_{t,i}=0\right\}.
\]
The oracle is non-deployable and is not a performance upper bound. It controls whether update points are benchmark-labeled normal while leaving the remaining adaptation procedure matched; it does not make the selected normal points maximally informative or the update rule optimal.

For $r\in\{P,O\}$, both adaptive settings apply the same online update rule,
\[
\theta^{r}_{t+1}
=
\operatorname{Update}
\!\left(
\theta^{r}_{t},
\nabla_{\theta}
\mathcal{L}_{\mathrm{TTA}}
(\theta^{r}_{t};X_t,m^{r}_{t})
\right),
\]
with the same source initialization, optimizer, learning rate, online order, updatable parameter set, and optimizer-step schedule. At each online step $t$, anomaly scores for $X_t$ are computed and recorded using the current parameters $\theta_t^r$ before any update on that window.
The update then produces $\theta_{t+1}^r$, which is used only for subsequent test windows; scores already recorded for $X_t$ are not recomputed after the update. Thus, the controlled difference between Practical and Oracle TTA is the update mask produced by the selector, which also changes the total contribution of selected points to the masked update loss.

We evaluate detection using the area under the precision--recall curve (AUPRC) without point adjustment. For multi-entity datasets, the main paired analysis uses a global common physical support. Let $\mathcal{E}_d$ denote the retained entities for dataset $d$, shared by Frozen, Practical, and Oracle across the main optimizer--seed grid. For adaptive setting $r\in\{P,O\}$, optimizer configuration $c$, and seed $s$, we define the paired detection effect as
\[
\Delta_{r,d,c,s}
=
\operatorname{median}_{e\in\mathcal{E}_d}
\left[
\mathrm{AUPRC}_{r,d,e,c,s}
-
\mathrm{AUPRC}_{F,d,e,s}
\right].
\]
For a single-entity dataset, this reduces to the corresponding paired
difference. Thus, physical entities are paired before dataset-level
aggregation; we do not subtract independently aggregated arm medians.

AUPRC is expressed on the $[0,1]$ scale, and $\Delta$ is the corresponding signed paired effect. When reporting effects in percentage points, we use $100\Delta$. We classify a cell
as showing a clear detection gain only when its full-precision paired effect satisfies $\Delta>0.005$, corresponding to more than $0.5$ percentage points.

Writing $\Delta_{\mathrm{prac}}=\Delta_P$ and
$\Delta_{\mathrm{orc}}=\Delta_O$, the paired comparison yields four
outcomes: both improve (BI), when both effects exceed $0.005$;
oracle-only gain (OOG), when
$\Delta_{\mathrm{orc}}>0.005$ and
$\Delta_{\mathrm{prac}}\leq0.005$;
neither improves (NI), when both effects are at most $0.005$; and
practical-only gain (POG), when
$\Delta_{\mathrm{prac}}>0.005$ and
$\Delta_{\mathrm{orc}}\leq0.005$.
NI denotes the absence of a clear gain under either adaptive setting; it includes near-zero changes as well as deterioration and does not imply that every NI cell degrades detection. An OOG outcome is consistent with the evaluated practical selector being limiting in that cell. An NI outcome indicates only that the failure persists under the tested benchmark-labeled normal-only  replacement; it does not identify which remaining component of the adaptation procedure is responsible.


\section{Experiments and Results}
\label{sec:5}

\paragraph{Experimental setup.}
The main study evaluates the AE backbone on nine benchmark families,
four adaptation optimizer configurations, and five random seeds,
yielding 180 dataset--optimizer--seed cells.
Within each cell, Frozen, Practical, and Oracle share the source checkpoint. Practical and Oracle use a matched online adaptation configuration, with the update mask produced by the selector as the only controlled difference. Detailed dataset and protocol settings are provided in Appendix~\ref{app:exp}.

\subsection{Oracle normal-only updates do not guarantee detection gains}
\label{sec:5_1}

We first examine whether benchmark-labeled normal-only adaptation reliably
improves over the corresponding frozen detector. Across the 180 matched
dataset--optimizer--seed cells, Oracle normal-only TTA produces a clear AUPRC
gain in 90 cells and no clear gain in the remaining 90. The median paired
Oracle-minus-Frozen AUPRC effect is 0.0033, with cell-level effects ranging
from $-0.2169$ to $+0.5840$. Thus, even when practical selection is
replaced by benchmark-labeled normal-only selection, adaptation does not yield
a uniformly beneficial detection effect.

\begin{table}[!ht]
\caption{
Oracle-minus-Frozen paired AUPRC effects by dataset. For each dataset, results are summarized over the 20 combinations of four optimizer configurations and five random seeds. For each combination, the paired effect is the median of the effects across physical entities in the common support. Displayed values are in percentage points.
}
\label{tab:5_1_1}
\centering
\small
\begin{tabular*}{0.9\linewidth}{@{\extracolsep{\fill}}l r r r r@{}}
\toprule
Dataset & Mean & Median & Minimum & Maximum \\
Daphnet & $-8.75$ & $-7.16$ & $-21.69$ & $-0.03$ \\
GECCO   & $+44.66$ & $+45.46$ & $+12.26$ & $+58.40$ \\
Genesis & $-0.21$ & $-0.04$ & $-0.72$ & $+0.00$ \\
MSL     & $-0.03$ & $-0.01$ & $-0.11$ & $+0.00$ \\
PSM     & $+12.35$ & $+13.46$ & $-2.29$ & $+24.70$ \\
PUMP    & $+32.76$ & $+31.91$ & $+24.99$ & $+41.27$ \\
SMAP    & $-0.03$ & $0.00$ & $-0.11$ & $0.00$ \\
SMD     & $+2.06$ & $+1.65$ & $+0.02$ & $+6.07$ \\
SWAT    & $+12.21$ & $+13.09$ & $+7.17$ & $+15.03$ \\
\bottomrule
\end{tabular*}
\end{table}

Table~\ref{tab:5_1_1} shows pronounced heterogeneity across datasets and
adaptation configurations. GECCO, PUMP, PSM, and SWAT have positive average
Oracle-minus-Frozen effects, whereas Daphnet is negative and Genesis, MSL,
and SMAP remain near zero. SMD shows a smaller positive effect. Overall,
benchmark-labeled normal-only adaptation can substantially improve detection
in some settings, while providing little gain or degrading detection in others.


Together, these results show that access to benchmark-labeled normal update
samples is not sufficient to guarantee successful TTA under the studied
reconstruction objective and matched update protocol. Even under
benchmark-labeled normal-only selection, detection effects remain heterogeneous
across datasets and adaptation configurations.

\subsection{Oracle normal-only TTA reveals outcomes hidden by practical TTA alone}
\label{sec:5_2}

\begin{figure}[ht]
    \centering
    \includegraphics[width=0.90\linewidth]{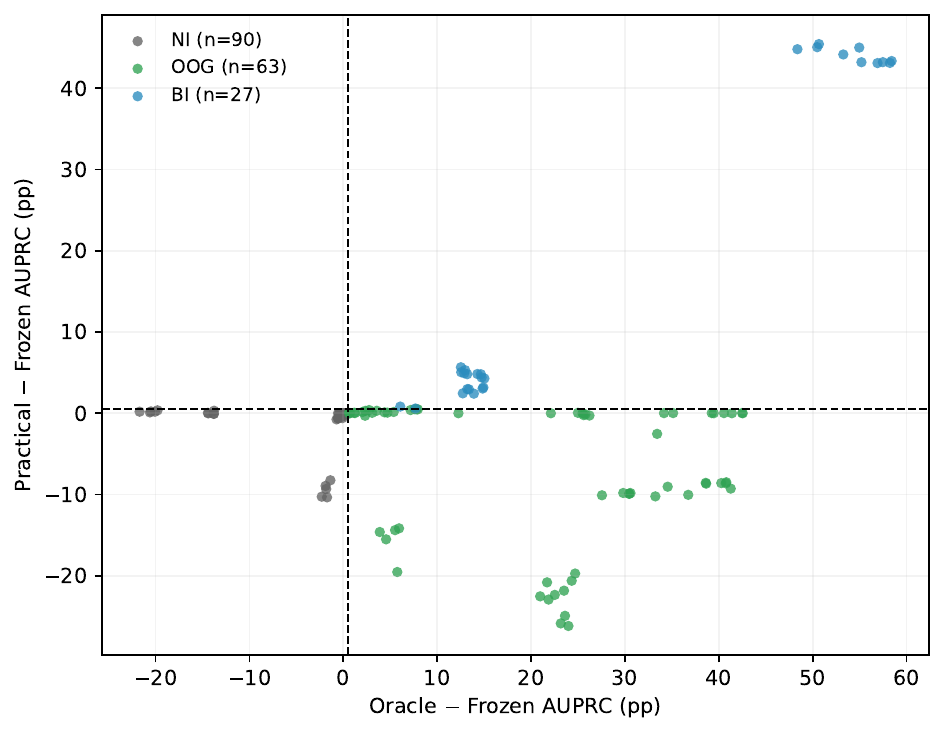}
    \caption{
    Matched-cell outcome categories from comparing practical TTA with oracle normal-only TTA.
    The figure plots paired cell-level gains over the frozen detector, with
    $\Delta_{\mathrm{orc}}$ on the x-axis and $\Delta_{\mathrm{prac}}$ on the y-axis.
    Dashed lines indicate the clear-gain threshold of $+0.5$ percentage points (pp). BI denotes a clear gain under both Practical and Oracle, OOG denotes a clear gain only under Oracle, NI denotes no clear gain under either setting, and POG denotes a clear gain only under Practical. No POG case occurs in the main grid.
    }
    \label{fig:5_2_1}
\end{figure}

Comparing only the frozen detector with practical TTA shows whether a deployed
TTA procedure helps, but it does not explain why it succeeds or fails.
We therefore compare Practical TTA with its matched Oracle normal-only
counterpart, differing in the update mask produced by the selector while keeping the
remaining online adaptation configuration matched. This comparison
distinguishes oracle-only-gain cases, where the benchmark-labeled normal-only
replacement recovers a detection gain, from neither-improves cases, where the
failure persists under the same replacement. An NI outcome does not identify the remaining failure mechanism; it establishes only that replacing the evaluated practical selector with benchmark-labeled normal-only selection is insufficient to produce a clear detection gain in that setting, motivating the objective-level analysis in the next subsection. We use the paired effects $\Delta_{\rm prac}$ and $\Delta_{\rm orc}$ and the four outcome categories defined above.

Figure~\ref{fig:5_2_1} shows that the matched comparison separates
practical-TTA failures into distinct outcomes. Out of 180 matched cells,
63 (35.0\%) are oracle-only-gain cases: Practical TTA does not obtain a clear
gain, whereas replacing its selector with benchmark-labeled normal-only
selection does. These cases are consistent with the evaluated practical
selector being limiting. In contrast, 90 cells (50.0\%) are neither-improves
cases, where neither Practical nor Oracle TTA obtains a clear gain over the
Frozen detector. These failures therefore persist under the tested
benchmark-labeled normal-only replacement. The remaining 27 cells (15.0\%)
improve under both Practical and Oracle TTA, and no cell falls into the
practical-only-gain category.

The exact outcome proportions depend on the practical adaptation variant, but the two qualitatively different failure regimes remain visible under substantially different selection and preprocessing choices. In the 45-setting robustness evaluation, the primary Static-P95/B1 configuration yields BI/OOG/NI/POG counts of 10/15/20/0. The corresponding counts are 9/21/15/0 for an M2N2-style exponential moving average (EMA) detrending variant and 13/12/20/0 for a CANDI-style False Positive Mining (FPM) whole-window selector. Thus, changing the practical selection or preprocessing mechanism changes the exact balance between oracle-rescuable and other outcomes, but both oracle-only-gain and neither-improves cases remain substantial.

Overall, the matched comparison reveals that practical-TTA failure is not a single regime. Some failures are oracle-rescuable and therefore consistent with the evaluated selector being limiting, whereas others persist under benchmark-labeled normal-only selection. The latter cases do not identify a unique remaining cause, but they show that improving sample selection alone is not a complete solution under the evaluated adaptation configurations. We next examine whether these persistent failures can be explained simply by failure to improve the reconstruction objective.

\subsection{Objective-level gains do not reliably imply detection gains}
\label{sec:5_3}

Sections~\ref{sec:5_1} and~\ref{sec:5_2} show that benchmark-labeled normal-only updates do not
reliably improve detection. We next ask whether these persistent failures simply
arise because adaptation fails to improve the reconstruction objective it
optimizes. Test-time adaptation minimizes reconstruction loss on selected
normal points, whereas anomaly detection depends on the relative ranking of
normal and anomalous scores over the full test stream. We focus on Oracle
normal-only TTA to examine this question after replacing the evaluated
practical selector with benchmark-labeled normal-only selection.

For each Oracle run, we measure objective improvement by the immediate
decrease in reconstruction loss induced by the online updates. At online step
$t$, let $\theta_t^O$ and $\theta_{t+1}^O$ denote the Oracle parameters
immediately before and after the update on $X_t$, respectively. For each
benchmark-labeled normal point contributing to that update, we define the
pre--post loss decrease as
\[
\delta_{t,i}^{O}
=
\ell\!\left(x_{t,i}, \hat{x}_{t,i}(\theta_t^O)\right)
-
\ell\!\left(x_{t,i}, \hat{x}_{t,i}(\theta_{t+1}^O)\right).
\]
Within each physical entity, we average $\delta_{t,i}^{O}$ over the
benchmark-labeled normal points contributing to Oracle updates. For
multi-entity datasets, we then take the median of the finite entity-level
values to obtain one objective-reduction value for each
dataset--optimizer--seed cell. Positive values indicate that the Oracle
updates decrease their immediate reconstruction objective. Detection improvement is measured as \(\Delta_{\rm orc}\), the effect of Oracle normal-only TTA relative to the Frozen detector. To understand when objective improvement does or does not translate into detection gain, we first examine the cell-level relationship and then analyze how adaptation reshapes difficult normal--anomaly score orderings.

\begin{figure}[ht]
\centering
\includegraphics[width=0.8\linewidth]{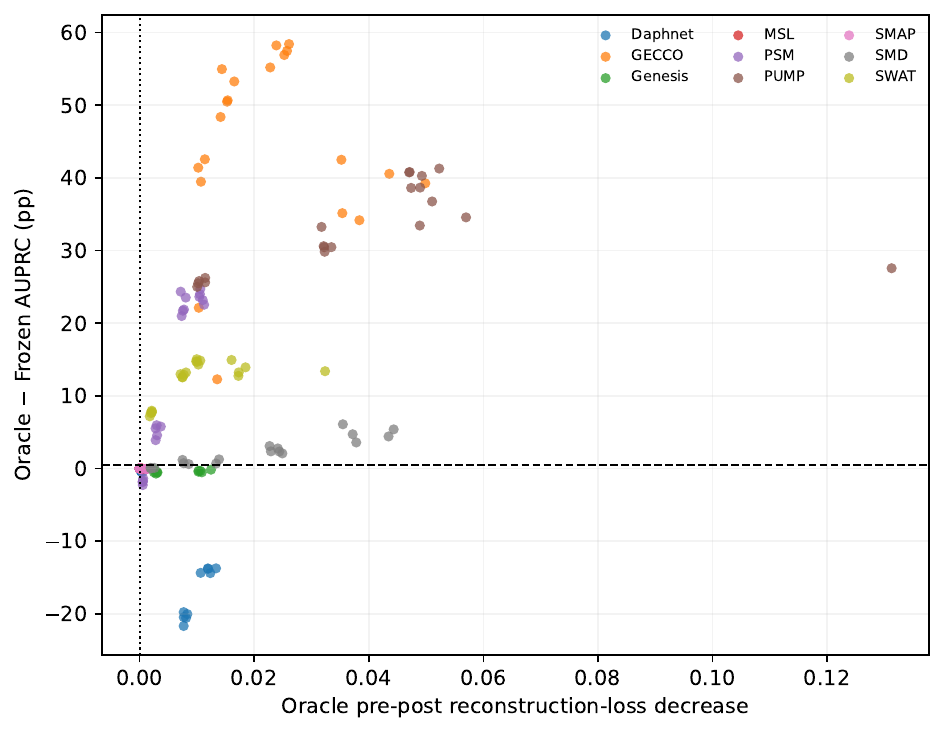}
\caption{
Objective-level improvement versus detection-level improvement under Oracle
normal-only TTA. Each point represents one dataset--optimizer--seed cell.
The x-axis reports the immediate pre--post decrease in reconstruction loss on
benchmark-labeled normal points contributing to Oracle updates; positive values
indicate that the online updates decrease their immediate reconstruction
objective. Objective decreases are averaged over benchmark-labeled normal update points
within each physical entity and then aggregated by the median over finite
entities; this diagnostic support can differ from the common physical support
used for the paired detection effect. The y-axis reports the paired Oracle-minus-Frozen AUPRC effect in percentage points (pp). The dotted vertical line marks zero objective decrease, and the
dashed horizontal line marks the +0.5 pp clear-gain threshold for AUPRC. Colors denote datasets.
}
\label{fig:5_3_1}
\end{figure}

Figure~\ref{fig:5_3_1} shows that the Oracle online updates decrease their immediate reconstruction objective across all 180 cells, but these objective-level improvements do not reliably imply detection success. All 180 objective reductions are positive, whereas 75 of the 180 Oracle-minus-Frozen detection effects are negative. Objective reduction and detection effect have a positive rank association (Spearman $\rho=0.605$), but its dataset-cluster 95\% confidence interval is wide ($[-0.020,\,0.811]$). Thus, reconstruction-objective improvement alone does not provide a reliable criterion for detection-level improvement.

Reducing reconstruction error on benchmark-labeled normal update samples does
not directly constrain how anomalous scores move relative to normal scores.
We therefore examine difficult parts of the score distribution: hard normals,
which have high Frozen anomaly scores, and boundary anomalies, which have low
Frozen anomaly scores. Hard normals are benchmark-normal points in the upper Frozen-score tail,
whereas boundary anomalies are anomalous points in the lower Frozen-score tail.
Selective compression measures whether Oracle reduces hard-normal scores more
than boundary-anomaly scores, while net pair gain measures the change in the fraction of correctly ordered pairs of hard normal and boundary anomaly points.

Across the full 180-cell grid, larger Oracle detection gains are associated
with more favorable score correction. The pooled Spearman correlations with
Oracle-minus-Frozen AUPRC are $\rho=0.736$ for selective compression and
$\rho=0.745$ for net pair gain. The same relationship is visible across the
nine dataset medians ($\rho=0.667$ and $0.728$, respectively). In contrast,
after ranking cells within each dataset and pooling those within-dataset ranks,
the associations are much weaker ($\rho=0.144$ and $0.106$). Thus, the
systematic relationship is primarily a between-dataset/regime pattern rather
than a universal optimizer- or seed-level law within each dataset. These
statistics are associative rather than causal and should not be interpreted
as necessary or sufficient conditions for adaptation success.

Appendix~\ref{app:score_shift_examples} provides representative visual examples: PUMP illustrates a setting with favorable hard-normal compression and a large detection gain, whereas Daphnet illustrates that favorable-looking local score shifts need not be sufficient for improved anomaly ranking.

\subsection{How robust are these findings?}
\label{sec:q4}

\paragraph{Cross-backbone robustness.}
We first examine whether the two failure regimes are specific to the main AE
backbone. Table~\ref{tab:backbone_robustness} extends the matched comparison
to USAD and LSTMED, with both full-model adaptation and separately audited
decoder-only variants where available. Both oracle-only-gain (OOG) and
neither-improves (NI) outcomes remain present in every evaluated backbone and
update scope. Thus, the coexistence of oracle-rescuable and persistent
failures is not confined to the main AE configuration.

\begin{table}[!htbp]
\centering
\small
\setlength{\tabcolsep}{4pt}
\caption{
Cross-backbone and update-scope outcome decomposition under matched
Practical--Oracle adaptation comparisons. Each row contains 180 logical
dataset--optimizer--seed cells. Entries report counts with row percentages
in parentheses. BI denotes cells where both Practical and Oracle show a clear gain; OOG denotes cells with no clear Practical gain but a clear Oracle gain; NI denotes cells with no clear gain under either setting; and POG denotes cells with a clear Practical gain but no clear Oracle gain.
A clear gain requires $\Delta>0.005$. Each row is aggregated on its own matched Frozen--Practical--Oracle physical support. LSTMED denotes the LSTM-based encoder--decoder backbone.
}
\label{tab:backbone_robustness}
\begin{tabular}{llccccc}
\toprule
Backbone & Update scope & Cells & BI & OOG & NI & POG \\
\midrule
AE
& Full model
& 180
& 27 (15.0\%)
& 63 (35.0\%)
& 90 (50.0\%)
& 0 (0.0\%) \\

USAD
& Full model
& 180
& 29 (16.1\%)
& 52 (28.9\%)
& 99 (55.0\%)
& 0 (0.0\%) \\

USAD
& Decoder only
& 180
& 33 (18.3\%)
& 42 (23.3\%)
& 105 (58.3\%)
& 0 (0.0\%) \\

LSTMED
& Full model
& 180
& 12 (6.7\%)
& 28 (15.6\%)
& 140 (77.8\%)
& 0 (0.0\%) \\

LSTMED
& Decoder only
& 180
& 8 (4.4\%)
& 26 (14.4\%)
& 146 (81.1\%)
& 0 (0.0\%) \\
\bottomrule
\end{tabular}
\end{table}

\paragraph{Alternative adaptation variants.}
Both oracle-only-gain (OOG) and neither-improves (NI) cases remain present
under the evaluated alternative adaptation variants. We also evaluate a
CANDI-style configuration combining False Positive Mining (FPM) with
Spatiotemporally-Aware Normality Adaptation (SANA), the two core components
of CANDI. In this setting, the AE backbone is frozen and only the SANA
input/output adaptation modules are updated. We compare Practical TTA using
FPM-selected windows with two Oracle variants: one updates on every window
whose benchmark labels are all normal, while the other retains the FPM
candidate set but removes windows containing benchmark-labeled anomalies.
These comparisons yield BI/OOG/NI/POG counts of 7/18/20/0 and 7/1/37/0,
respectively.

The two regimes are also substantially stable under perturbations of the
benchmark labels. We consider entity-wise, segment-relative 10\% boundary
dilation and erosion, together with a fixed 10\% false-negative contamination
realization. Each perturbation is applied either to the Oracle update mask
(mask-only), to the evaluation labels for all three arms (evaluation-only), or
to both (joint). Boundary dilation preserves the outcome category of all
45/45 cells in all three channels. Boundary erosion preserves 45/45 categories
in the mask-only channel and 41/45 in each of the evaluation-only and joint
channels. False-negative contamination preserves 40/45, 44/45, and 39/45
categories in the three channels, respectively.

Changing the online update cadence likewise retains both regimes. For cadence
$B\in\{1,2,4,8\}$, every chronological window is scored and contributes its
masked loss, while gradients are accumulated over blocks of $B$ consecutive
windows before each optimizer step. The corresponding BI/OOG/NI/POG counts for
B1, B2, B4, and B8 are 10/15/20/0, 11/14/20/0, 12/13/20/0, and 7/13/25/0,
respectively. Thus, the coexistence of oracle-rescuable and persistent failures
is not restricted to the primary update cadence.

Finally, uncertainty estimates reflect substantial heterogeneity across the
nine benchmark families. A 10,000-replicate dataset-cluster bootstrap gives
an OOG proportion of $0.35$ with a 95\% confidence interval of
$[0.122,\,0.600]$, and an NI proportion of $0.50$ with a 95\% confidence
interval of $[0.194,\,0.806]$.
These intervals emphasize that the 180 optimizer--seed cells should not be
interpreted as 180 independent dataset-level observations.

Overall, these checks change the exact outcome proportions but preserve both
oracle-rescuable and persistent failures.
\section{Discussion and Conclusion}

We studied test-time adaptation for reconstruction-based time-series anomaly detection under deployment-time distribution shift, focusing on the role of update-sample selection. Our study establishes three main empirical insights into reconstruction-based TSAD TTA. First, replacing a practical update selector with benchmark-labeled normal-only selection does not guarantee improved anomaly detection under the same matched adaptation configuration. Second, paired practical and normal-only adaptation reveals qualitatively different failure regimes: some practical failures are rescued by the normal-only intervention, whereas others persist even after this replacement. Third, improving the reconstruction objective on benchmark-labeled normal update samples does not reliably translate into improved anomaly ranking. Our score-level analysis further shows that larger detection gains under normal-only adaptation are associated across datasets and adaptation regimes with more favorable correction of difficult normal--anomaly score orderings, although this relationship is primarily a between-dataset/regime pattern rather than a universal within-dataset law.

Together, these findings show that update-sample selection matters, but changing selection alone is not sufficient to ensure detection gains in the adaptation settings studied here. When replacing practical selection with benchmark-labeled normal-only selection recovers detection gains, the result is consistent with the evaluated selector being limiting. When failure persists under the same intervention, however, the normal-only replacement alone is insufficient to recover detection improvement, motivating examination of the broader adaptation mechanism, including the update objective, optimization dynamics, update scope, model behavior, and their interaction with anomaly-score distributions. The matched normal-only intervention should therefore be interpreted as a diagnostic tool for studying the role of update-sample selection, rather than as a deployable TTA method or a performance upper bound.

Our conclusions are deliberately scoped to the evaluated reconstruction-based TSAD settings with online parameter updates. Forecasting-based, density-based, or other adaptation paradigms may use different objectives, score semantics, and update mechanisms, so the empirical findings here should not be assumed to transfer directly to those settings. In addition, the normal-only intervention is conditioned on benchmark labels. If those labels miss subtle anomalies, conflate benign shifts with anomalies, or reflect dataset-specific annotation conventions, the resulting update mask inherits these limitations. Thus, the intervention provides access to benchmark-labeled normal samples, not to a perfect semantic notion of normality.

Overall, understanding reconstruction-based TSAD TTA requires considering not only which samples
are used for updating, but also how the resulting updates reshape the relative scores of normal and
anomalous observations.

\newpage
\subsection*{AI use statement}
Large language models and AI-assisted coding tools were used to support research ideation and execution, including experimental planning, code development and debugging, analysis planning, and interpretation of results. They were also used to assist with drafting and polishing portions of the manuscript. The authors take responsibility for the final content of this work, including its experimental designs, implementations, analyses, scientific claims, citations, and manuscript text.

\subsection*{Reproducibility statement}
We specify the datasets, preprocessing and window settings, optimizer configurations, random seeds, online update protocol, evaluation metrics, and aggregation procedures used in our experiments. Appendix~\ref{app:exp} further documents the compute resources, common support and paired aggregation, and matched practical/oracle adaptation protocol. We will release the implementation, experiment configurations, and result-processing scripts upon publication.

\bibliographystyle{iclr2027_conference}
\bibliography{references}

\ifshowappendix
\appendix
\counterwithin{table}{section}
\counterwithin{figure}{section}
\renewcommand{\thetable}{\thesection.\arabic{table}}
\renewcommand{\thefigure}{\thesection.\arabic{figure}}

\newpage
\section{Experimental and Protocol Details}
\label{app:exp}
\setcounter{table}{0}
\setcounter{figure}{0}

\subsection{Dataset Details}
\label{app:dataset}

We evaluate this oracle-based comparison on nine TSAD benchmarks: 
Daphnet~\citep{Daphnet},
MSL and SMAP~\citep{MSL_SMAP},
PSM~\citep{PSM},
PUMP~\citep{PUMP},
SMD~\citep{SMD},
GECCO~\citep{GECCO},
Genesis~\citep{Genesis},
and SWAT~\citep{SWAT}. These datasets span diverse application
domains, including water treatment, server monitoring, spacecraft telemetry,
health monitoring, and industrial or cyber-physical monitoring. Together, they
provide a heterogeneous testbed for studying test-time adaptation under
train--test distribution shift. Summary statistics of the datasets used in our
experiments are provided in Table~\ref{tab:dataset-properties}.

For multi-entity benchmarks such as SMD, MSL, and SMAP, the reported lengths, anomaly ratios, and main results are aggregated at the benchmark level unless otherwise stated. In Table~\ref{tab:dataset-properties}, \textit{Obtained from} indicates the repository or benchmark package from which the data were obtained for our experiments, rather than the original dataset paper. This distinction is important because some datasets were obtained from their original public repositories, whereas others were obtained from benchmark packages. When needed, we analyze variation across subsets separately in additional experiments, rather than using subsets as the default reporting unit in the main evaluation.

Table~\ref{tab:dataset-properties} reports benchmark-level channel counts. In the model description,
however, $D$ denotes the processed feature count of the current physical
entity. Because preprocessing may remove features and multi-entity benchmarks
may have entity-specific feature dimensions, $D$ need not equal the
benchmark-level channel count reported in Table~\ref{tab:dataset-properties}.

The reported train and test lengths refer to the full processed splits before
incomplete-window truncation. Test anomaly ratios are computed from the
processed point-level test labels; for multi-entity benchmarks, anomalous and
total test points are pooled across entities before computing the ratio.

\begin{table}[h]
\footnotesize
\centering
\caption{
Dataset properties used in the main experiments.
\textit{Obtained from} indicates the repository or benchmark package from
which the data were obtained for our experiments.
}
\label{tab:dataset-properties}
\setlength{\tabcolsep}{3pt}
\begin{tabularx}{\columnwidth}{lXrrrrrX}
\toprule
Dataset
& Domain
& \makecell[c]{Entities/\\Subsets}
& Channels
& Train Len
& Test Len
& \makecell[c]{Test Anomaly\\Ratio (\%)}
& Obtained from \\
\midrule
SWAT
& Water treatment
& 1
& 51
& 495,000
& 449,919
& 12.14
& TAB \\

SMD
& Server machine
& 28
& 38
& 708,405
& 708,420
& 4.16
& OmniAnomaly \\

MSL
& Spacecraft
& 27
& 55
& 58,317
& 73,729
& 10.48
& Telemanom \\

SMAP
& Spacecraft
& 55
& 25
& 140,825
& 444,035
& 12.83
& Telemanom \\

PSM
& Server machine
& 1
& 25
& 132,481
& 87,841
& 27.76
& RANSynCoders \\

Daphnet
& Health
& 1
& 9
& 9,693
& 29,081
& 7.93
& TSB-AD \\

PUMP
& Water treatment
& 1
& 44
& 110,151
& 110,151
& 9.40
& TAB \\

GECCO
& Water treatment
& 1
& 9
& 16,165
& 122,356
& 1.41
& TSB-AD \\

Genesis
& Machinery
& 1
& 18
& 4,055
& 12,165
& 0.41
& TSB-AD \\
\bottomrule
\end{tabularx}
\end{table}

\paragraph{Implementation basis and data sources.}
Our implementation is built by modifying the open-source Time-Series-Library codebase, which is released under the MIT License. For benchmark sources, TSB-AD is released under Apache-2.0, OmniAnomaly under MIT, Telemanom under a BSD-style Caltech/JPL license, and RANSynCoders under a BSD License. For TAB, we did not find an explicit repository license in the GitHub repository at access time; we therefore report it as an acquisition source rather than assigning a license ourselves. The benchmark datasets used in our experiments were obtained from the corresponding sources listed in Table~\ref{tab:dataset-properties}.

\subsection{Windowing Protocol}
\label{app:dataset_protocol}

Across all nine benchmark families, we use a window size of 12 and a test
stride of 12. Test windows are therefore non-overlapping and are processed
sequentially in chronological order during online adaptation.

\subsection{Backbone and Optimization Details}

The main results in the paper are based on a reconstruction-based AE backbone. The model is first trained on the training split and then evaluated under the frozen, practical TTA, and oracle normal-only TTA settings using the same pre-trained source model. Reconstruction losses and anomaly scores are computed in standardized feature
space, using the feature-wise mean squared reconstruction error defined in
Section~\ref{sec:setup}; no inverse transformation is applied before scoring.

\paragraph{AE architecture.}
A batch of input windows has shape \([B, 12, D]\), where \(B\) is the batch size and \(D\) is the processed feature count of the current physical entity. Each window is flattened to \(F = 12D\) features before encoding. The encoder has dimensions
\[
F \rightarrow H_1 \rightarrow H_2 \rightarrow L,
\]
where
\[
H_1=\max(1,\lfloor F/2\rfloor),\qquad
H_2=\max(1,\lfloor F/4\rfloor),\qquad
L=\min(64,\max(8,\lfloor F/8\rfloor)).
\]
A ReLU follows each encoder linear layer. The decoder reverses these
dimensions as
\[
L \rightarrow H_2 \rightarrow H_1 \rightarrow F,
\]
with ReLU activations after its first two linear layers and no output
activation, and reshapes the reconstruction back to $[B,12,D]$. During the
main full-model TTA protocol, all encoder and decoder parameters are updated;
Frozen, Practical, and Oracle use the same entity- and seed-specific
source-trained checkpoint.

For adaptation, we evaluate four optimizer configurations: SGD with learning rates $10^{-3}$ and $10^{-4}$, and AdamW with learning rates $10^{-4}$ and $3\times 10^{-4}$. These configurations are applied consistently across practical and oracle TTA within each optimizer setting, so practical-vs-oracle comparisons are not confounded by different updater choices. Comparisons against the frozen detector still measure the combined effect of online updating and the selected update mask. All experiments are repeated over five random seeds. The main analysis uses
the resulting dataset--optimizer--seed cells to compute paired Practical-minus-Frozen
and Oracle-minus-Frozen detection effects as defined in Section~\ref{sec:setup}.

We report the backbone and optimization settings needed to interpret the comparisons reported in the main text.

\paragraph{Common support and paired aggregation.}
For each multi-entity dataset, the main analysis uses one global retained
physical-entity support shared by Frozen, Practical, and Oracle across the
optimizer--seed grid. Paired detection effects are first computed at the
physical-entity level as Practical-minus-Frozen and Oracle-minus-Frozen
differences, and are then aggregated by the median within each
dataset--optimizer--seed cell. For single-entity datasets, this reduces to the
corresponding paired difference.

\subsection{Protocol Implementation Details}

The frozen, practical TTA, and oracle normal-only TTA settings all start from the same source checkpoint for each physical entity and seed. The frozen detector
performs no test-time parameter updates. Practical TTA and Oracle normal-only TTA use the matched update protocol: the same model initialization, updatable parameter set, optimizer family,
learning rate, update rule, and online order.

Practical TTA uses the score-thresholded predicted-normal selector defined in
Section~\ref{sec:setup}. For each physical entity and seed, the official source-training
series is split chronologically: the first 80\% is used to train the source
model and the final 20\% is used as the source-validation split. The frozen
source checkpoint is selected by minimum reconstruction loss on this validation
split. Using that checkpoint, we compute point-level reconstruction scores on
non-overlapping length-12 validation windows with stride 12, discarding an
incomplete final window, and set $\tau_{\mathrm{sel}}$ to the 95th percentile
of these scores. Calibration is performed separately for each physical entity
and seed; multi-entity datasets therefore use entity-specific thresholds.
No test data or test labels are used for calibration.

During online testing, $\tau_{\mathrm{sel}}$ remains fixed. For each arriving
test window, Practical TTA computes point-level reconstruction scores using the
current pre-update model and selects points whose scores are strictly below
$\tau_{\mathrm{sel}}$. The threshold is not recalibrated from past or future
test scores.

For both Practical and Oracle TTA, the point-level anomaly scores retained for
detection evaluation are the scores computed before the update on the current
window. After these scores are recorded, the corresponding update mask is
constructed and the online parameter update is applied. The updated parameters
are first used when processing the next test window; the current window is not
rescored after adaptation.

Oracle normal-only TTA keeps the matched update protocol but replaces the practical predicted-normal mask with a benchmark-label-conditioned normal mask. For each arriving test window, only benchmark-labeled normal points contribute to the reconstruction update loss; all other point losses are set to zero before averaging over the full window. If the window contains no benchmark-labeled normal points, its masked data loss is zero, but the scheduled optimizer step is still executed; retained optimizer state can therefore change the parameters. Benchmark labels are used only for this diagnostic Oracle update selection and for final evaluation; they are not used to tune the threshold, learning rate, optimizer choice, update scope, or any other hyperparameter.

This implementation keeps the adaptation procedure matched between Practical
and Oracle TTA except for the update mask. The optimizer, parameter update
rule, learning rate, online order, update schedule, and source initialization
are shared. Thus, the comparison isolates the effect of replacing the
Practical predicted-normal selector with benchmark-normal selection, including
the resulting change in how much selected data contributes to the update loss.

\subsection{Cross-Backbone and Update-Scope Robustness}
\label{app:cross_backbone}
\paragraph{Cross-backbone protocol.}
We additionally evaluate USAD, a dual-decoder reconstruction model, and
LSTMED, a recurrent encoder--decoder reconstruction model. Each reported
backbone--scope setting uses the same nine benchmark families, four optimizer
configurations, and five random seeds as the main AE study, yielding 180
dataset--optimizer--seed cells. The Practical and Oracle comparisons follow
the same intervention semantics as in Section~3: Practical selection is
calibrated from the corresponding frozen source model, whereas Oracle selection
uses benchmark-normal labels. Detection scores are recorded before each online
update, and outcome categories use the same $\Delta>0.005$ clear-gain
criterion.

\paragraph{Update scope.}
Full-model adaptation updates all trainable detector parameters. Decoder-only
adaptation freezes the encoder and updates only decoder-side reconstruction
parameters. For USAD, this freezes the shared encoder and updates the decoder
modules; for LSTMED, it freezes the recurrent encoder and updates the
corresponding reconstruction decoder. Within each backbone--scope setting,
Practical and Oracle TTA use the same parameter scope, optimizer configuration,
online order, scoring procedure, and evaluation protocol; their controlled
difference remains the update-sample selection rule.

\paragraph{Windowing and matched support.}
The reported USAD and LSTMED studies use the same non-overlapping test
windowing protocol as the main analysis, with window length 12 and stride 12.
Within each backbone and update scope, Frozen, Practical, and Oracle are
evaluated on matched physical-entity support before aggregation to
dataset--optimizer--seed cells. All five reported backbone--scope settings
retain all 180 logical cells, although the retained physical-entity support
need not be identical across backbone or update-scope settings.

\subsection{Compute Resources}

All experiments were run on NVIDIA V100 GPUs with 32GB memory, with each run using a single GPU. The main AE study covers nine datasets, four optimizer configurations, and five random seeds, yielding 180 dataset--optimizer--seed cells. Additional selector, cadence, label-perturbation, and score-level analyses use the analysis-specific evaluation grids reported in the main text and appendix.

\FloatBarrier
\section{Additional Outcome and Robustness Results}
\label{app:robustness}
\setcounter{table}{0}
\setcounter{figure}{0}
\FloatBarrier

This appendix summarizes the outcome and robustness results supporting the
main analysis. Table~\ref{tab:main_grid_outcomes} reports the canonical
180-cell Practical--Oracle outcome counts,
Table~\ref{tab:robustness_variants} summarizes the evaluated selector,
adaptation, and cadence variants, and
Table~\ref{tab:label_perturbation} reports outcome-category retention under
benchmark-label perturbations. These results complement the corresponding
analyses in Sections~\ref{sec:5_2} and~\ref{sec:q4}.

\begin{table}[t]
\centering
\caption{
Main-grid outcome counts for the matched Practical--Oracle comparison.
The 180 cells correspond to nine datasets, four optimizer configurations,
and five random seeds. BI: both improve; OOG: oracle-only gain;
NI: neither improves; POG: practical-only gain. A clear gain means a paired AUPRC improvement over Frozen greater than $0.005$ (0.5 percentage points).
}
\label{tab:main_grid_outcomes}
\small
\begin{tabular}{lrrrrr}
\toprule
Analysis & Cells & BI & OOG & NI & POG \\
\midrule
Main AE grid & 180 & 27 & 63 & 90 & 0 \\
\bottomrule
\end{tabular}
\end{table}

Unless otherwise stated, the robustness analyses use the same 45 logical
dataset--optimizer--seed cells: the nine benchmark families, AdamW with
learning rate $10^{-4}$, and seeds 42--46. Static-P95/B1 is the corresponding
45-cell slice of the main 180-cell AE experiment and reuses its Frozen,
Practical, and Oracle runs. The robustness analyses share these logical cell
identities, although branch-specific analyses may use different matched
physical-entity support.

For cadence $B\in\{1,2,4,8\}$, every chronological test window is scored and masked, and its masked loss is scaled by $1/B$ and backpropagated. Gradients are
accumulated over each complete block of $B$ windows and an optimizer step is
taken at the end of the block. With zero-based indexing, steps therefore occur
when $(t+1)\bmod B=0$. Model and optimizer state are retained between block
boundaries, and gradients from an incomplete final block are discarded without
an optimizer step. Practical and Oracle TTA use the same cadence schedule.

\paragraph{Robustness variants.}
Static-P95/B1 is the AdamW-$10^{-4}$ slice of the main protocol and uses the
point-level source-validation P95 selector with full-AE updates on every test
window. The M2N2-style variant maintains an online exponential moving mean,
reconstructs the detrended current window, restores the mean before scoring,
and uses the same P95 rule for Practical updates; its matched baseline and
Oracle use the same detrending procedure. Both adaptive arms update the full AE.

\paragraph{CANDI-style FPM selector.}
False Positive Mining (FPM) is one of the two core components of CANDI
\citep{CANDI}. Our FPM-only variant transplants its score- and latent-similarity
based sample-selection principle into the main full-AE adaptation protocol.
The Practical point-level Static-P95 selector is replaced by a label-free
whole-window selector constructed from source-validation reconstruction-score
strata and proximity to frozen source-encoder latent references. Selected
windows update the full AE, while the Frozen and benchmark-normal point Oracle
controls are retained from the core protocol.

\paragraph{CANDI-style FPM+SANA variants.}
Spatiotemporally-Aware Normality Adaptation (SANA) is the second core component
of CANDI. In our FPM+SANA variants, the source AE is frozen and only the
input/output SANA modules are trainable. Practical TTA updates these modules on
FPM-selected windows. One Oracle variant updates on every window whose
benchmark labels are all normal, whereas the other preserves the FPM candidate
set and removes candidates containing any benchmark-labeled anomaly.

\begin{table}[!htbp]
\centering
\small
\setlength{\tabcolsep}{4pt}
\renewcommand{\arraystretch}{1.05}
\caption{
    Outcome counts for selector, adaptation, and cadence variants over 45 matched
    cells per analysis, comprising nine datasets, AdamW at learning rate
    $10^{-4}$, and seeds 42--46. BI denotes both improve, OOG denotes Oracle-only
    gain, NI denotes neither improves, and POG denotes Practical-only gain; a
    clear gain requires a paired AUPRC improvement over Frozen greater than
    $0.005$ (0.5 percentage points). For cadence $B$, gradients are accumulated
    over blocks of $B$ consecutive test windows before each optimizer step.
    Physical-entity support may differ across variants. EMA denotes exponential
    moving average. FPM denotes False Positive Mining and SANA denotes
    Spatiotemporally-Aware Normality Adaptation, the two evaluated CANDI
    components.
}
\label{tab:robustness_variants}
\begin{tabular}{p{0.55\linewidth}rrrr}
\toprule
Analysis & BI & OOG & NI & POG \\
\midrule
Static-P95 / B1
    & 10 & 15 & 20 & 0 \\

M2N2-style EMA detrending
    & 9 & 21 & 15 & 0 \\

CANDI-style FPM-only
    & 13 & 12 & 20 & 0 \\

FPM+SANA / all-normal-window Oracle
    & 7 & 18 & 20 & 0 \\

FPM+SANA / label-cleaned FPM Oracle
    & 7 & 1 & 37 & 0 \\

Cadence B2
    & 11 & 14 & 20 & 0 \\

Cadence B4
    & 12 & 13 & 20 & 0 \\

Cadence B8
    & 7 & 13 & 25 & 0 \\
\bottomrule
\end{tabular}
\end{table}
\FloatBarrier

Across these selector, adaptation, and cadence variants, the exact outcome
proportions change, but both OOG and NI remain present throughout the evaluated
settings.

\paragraph{Benchmark-label perturbations.}
Perturbations are constructed independently for each physical entity from its
point-level benchmark labels. An anomaly segment is a maximal contiguous run
of anomalous points in the full test sequence, independent of window
boundaries. For a segment of length $L$, the boundary width used for both
dilation and erosion is
\[
w(L)=
\begin{cases}
0, & L\leq 1,\\
\min\!\left\{
\max\!\left(1,\left\lceil 0.1L\right\rceil\right),
\left\lfloor\frac{L-1}{2}\right\rfloor
\right\}, & L>1.
\end{cases}
\]
Dilation relabels up to \(w(L)\) adjacent normal points on each side of the segment as anomalous,
stopping at the sequence boundaries. Erosion relabels the first and last \(w(L)\) anomalous points
of each segment as normal. The cap prevents complete removal of a segment; segments of length one
or two are unchanged.

For false-negative contamination, we relabel
\[
K = \left\lceil 0.1 N_{\mathrm{anom}} \right\rceil
\]
anomalous points as normal within each entity, where
\(N_{\mathrm{anom}}\) is the number of benchmark-labeled anomalous points in that entity. Using a fixed NumPy random seed of 20260725, we randomly permute anomaly segments and consume them in that order: whole segments are relabeled while they fit within the remaining budget, and,
when necessary, a contiguous block is sampled uniformly from the final partially used segment. We use one fixed realization per entity label sequence, reused across model seeds and perturbation channels.

In the mask-only channel, the perturbed labels modify only the Oracle update
mask, while evaluation uses the original benchmark labels. In the evaluation-only channel, Oracle update selection uses the original benchmark labels, while Frozen, Practical, and Oracle are all evaluated against the same perturbed label vector. In the joint channel, the same perturbed vector is used both to define the Oracle update mask and to evaluate all three arms.

\begin{table}[!htbp]
\centering
\small
\caption{
Outcome-category retention under benchmark-label perturbations on the same
45-cell AdamW-$10^{-4}$, seed-42--46 robustness grid. Boundary dilation and
erosion use segment-relative 10\% perturbations. 
For false-negative contamination, anomalous points are relabeled as normal using a 10\% anomaly-point budget and the fixed realization generated with seed 20260725. Each entry reports the number of cells whose outcome category matches the corresponding unperturbed analysis. Retention denotes
category agreement, not sample availability. Mask-only perturbs only the
Oracle update mask; evaluation-only perturbs the evaluation labels for all
three arms; joint applies the same perturbed labels to both the Oracle update
mask and the evaluation of all three arms.
}
\label{tab:label_perturbation}
\begin{tabular}{lccc}
\toprule
Perturbation & Mask-only & Evaluation-only & Joint \\
\midrule
Boundary dilation
    & 45/45 & 45/45 & 45/45 \\
Boundary erosion
    & 45/45 & 41/45 & 41/45 \\
False-negative contamination
    & 40/45 & 44/45 & 39/45 \\
\bottomrule
\end{tabular}
\end{table}

\FloatBarrier

\paragraph{Dataset-cluster uncertainty.}
To characterize uncertainty without treating the 180 optimizer--seed cells as
independent dataset-level observations, we use a dataset-cluster bootstrap over
the nine benchmark families. We draw 10,000 bootstrap replicates by resampling
datasets as clusters and recompute the corresponding outcome proportions within
each replicate. This procedure gives an OOG proportion of $0.35$ with a 95\%
confidence interval of $[0.122,\,0.600]$ and an NI proportion of $0.50$ with a
95\% confidence interval of $[0.194,\,0.806]$.

\FloatBarrier
\section{Additional Results for Section~\ref{sec:5_3}: Score-Shift Examples}
\label{app:score_shift_examples}
\setcounter{table}{0}
\setcounter{figure}{0}

\paragraph{Objective--detection association.}
Across the 180 main-grid cells, the immediate pre--post reconstruction-loss
decrease on benchmark-labeled normal points contributing to Oracle updates has
a Spearman rank correlation of $\rho = 0.605$ with the
Oracle-minus-Frozen AUPRC effect. For each physical entity, we average the
point-level loss decreases from immediately before to immediately after Oracle
updates over the benchmark-labeled normal points contributing to those updates,
and then take the median of the finite entity-level values within each
dataset--optimizer--seed cell.

The objective diagnostic uses the finite physical-entity support on which the
pre--post loss quantities are defined, whereas the paired detection effect uses
the canonical common physical support defined in Section~\ref{sec:setup}.
To reflect dependence among optimizer--seed cells from the same benchmark
family, uncertainty is estimated by clustering at the dataset level. The
resulting 95\% confidence interval is $[-0.020, 0.811]$. Thus, although the
point estimate is positive, immediate reconstruction-objective improvement
alone is not a reliable criterion for detection improvement.

\paragraph{Full-grid score-shift association.}
For each physical entity $e$, let
\[
\mathcal{N}_e=\{i:y_i=0\},
\qquad
\mathcal{A}_e=\{j:y_j=1\}.
\]
Under the primary tail setting, hard normals and boundary anomalies are defined
from Frozen scores as
\[
\mathcal{H}_e
=
\left\{
i\in\mathcal{N}_e:
s_i^F \ge Q_{0.85}(s_{\mathcal{N}_e}^F)
\right\},
\]
and
\[
\mathcal{B}_e
=
\left\{
j\in\mathcal{A}_e:
s_j^F \le Q_{0.30}(s_{\mathcal{A}_e}^F)
\right\}.
\]
Tail membership therefore depends only on benchmark labels and Frozen scores.
Quantile thresholds use linear interpolation, and points equal to a threshold
are included.

Let
\[
d_i=s_i^F-s_i^O
\]
denote the Frozen-to-Oracle score decrease. We define selective compression as
\[
\mathrm{SC}_e
=
\operatorname*{median}_{i\in\mathcal{H}_e} d_i
-
\operatorname*{median}_{j\in\mathcal{B}_e} d_j.
\]
Thus, $\mathrm{SC}_e>0$ means that hard-normal scores are preferentially
reduced relative to boundary-anomaly scores.

For each pair $(i,j)\in\mathcal{H}_e\times\mathcal{B}_e$, where $i$ is normal
and $j$ is anomalous, define
\[
C_{ij}^{M}
=
\mathbf{1}\!\left[s_j^{M}>s_i^{M}\right],
\qquad
M\in\{F,O\}.
\]
Score ties are treated as incorrectly ordered. Net pair gain is
\[
\mathrm{NPG}_e
=
\frac{1}{|\mathcal{H}_e||\mathcal{B}_e|}
\sum_{i\in\mathcal{H}_e}
\sum_{j\in\mathcal{B}_e}
\left(C_{ij}^{O}-C_{ij}^{F}\right).
\]
It therefore measures the change in the fraction of pairs of hard normal and
boundary anomaly points for which the anomalous point is ranked above the
normal point. Positive values indicate more wrong-to-correct than correct-to-wrong pair
transitions after Oracle adaptation.

Both statistics are computed independently within each physical entity. Net
pair gain uses the exact Cartesian product of the selected tails, without pair
sampling. For each dataset--optimizer--seed cell, entity-level statistics are
aggregated by the median over finite matched Frozen--Oracle entities. The
finite entity support used for the score-shift statistics is not always
identical to the canonical common support used for the paired detection effect,
because entities with non-finite score arrays are excluded from the score-shift
analysis.

Using the primary 15\%/30\% tail setting, we evaluate the association between
Oracle-minus-Frozen AUPRC and two score-ordering statistics across the 180
dataset--optimizer--seed cells. The pooled Spearman correlations are
$\rho=0.736$ for selective compression and $\rho=0.745$ for net pair gain.
Across the nine dataset medians, the corresponding correlations are
$\rho=0.667$ and $0.728$. After ranking the detection effects and score-shift
statistics within each dataset and pooling those within-dataset ranks, the
associations decrease to $\rho=0.144$ and $0.106$, respectively. These results
support a primarily between-dataset/regime association rather than a universal
within-dataset optimizer--seed relationship, and should be interpreted as
associative rather than causal.

\begin{figure}[t]
    \centering
    \includegraphics[width=\linewidth]{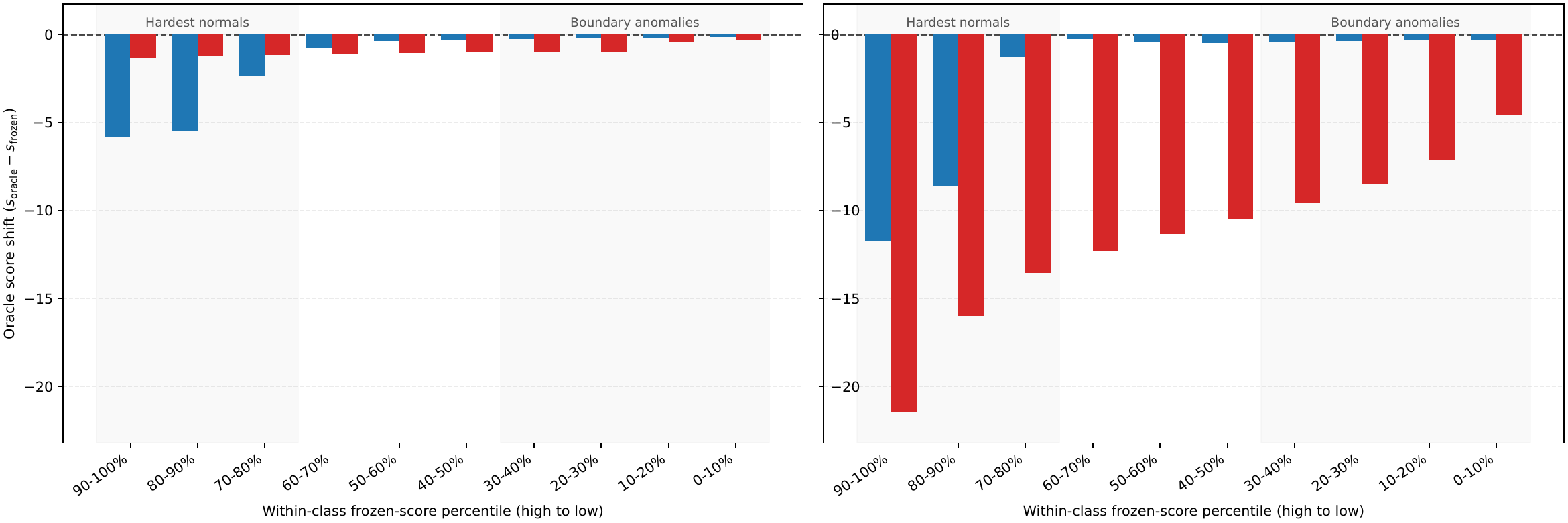}
    \caption{
        Representative Oracle normal-only score-shift cases for the AE backbone with AdamW at learning rate $10^{-4}$ and seeds 42--46. Within each seed, we compute the median score shift $s_{\mathrm{Oracle}}-s_{\mathrm{Frozen}}$ in class-specific Frozen-score decile bins, and each bar reports the median of these values across seeds.
        Blue bars denote normal points and red bars denote anomalous points; more
        negative values indicate stronger score reduction after Oracle normal-only
        adaptation. PUMP (left) illustrates a gain case, where high-scoring normal
        points are strongly compressed while low-scoring boundary anomalies are
        largely preserved. Daphnet (right) illustrates a failure case, where anomalous scores also decrease and detection worsens despite reconstruction-objective improvement. These illustrative decile bins are distinct from the 15\%/30\% tail definition used for the systematic selective-compression and net-pair-gain analysis.
    }
    \label{fig:score_shift_examples}
\end{figure}

Figure~\ref{fig:score_shift_examples} provides representative score-shift
examples for the objective--detection mismatch discussed in Section~\ref{sec:5_3}.
The examples are illustrative rather than statistical. They show how oracle
normal-only updates can reduce reconstruction loss while producing different
effects on anomaly ranking. In PUMP, high-scoring normal points show larger score reductions than low-scoring boundary anomalies, alongside improved detection. In Daphnet, anomalous scores also decrease substantially, while detection worsens despite reconstruction-objective improvement.

\fi
\end{document}